%% file: 1_main.tex
\documentclass{article}
\input{0_ieee.tex}
\input{0_preamble}

\begin{document}

	\title{To Be Defined: Knowledge Distillation and Object Detection}
    \title{Leveraging knowledge distillation for partial multi-task learning from multiple remote sensing datasets}
    \name{Hoàng-Ân Lê and Minh-Tan Pham~\thanks{This work was supported by the SAD 2021 ROMMEO project (ID 21007759) and the ANR AI chair OTTOPIA project (ANR-20-CHIA-0030).}}
    \address{IRISA, Université Bretagne Sud, UMR 6074, 56000 Vannes, France \\
    \tt\small \{hoang-an.le,minh-tan.pham\}@irisa.fr}
	\maketitle
	
\begin{abstract}
   \input{Sections/0_abstract}

\end{abstract}

\begin{keywords}
    Deep learning, multi-task learning,
    knowledge distillation, object detection, semantic segmentation, remote sensing
\end{keywords}

\input{Sections/1_Intro}

\input{Sections/3_Method}

\input{Sections/5_Experiments}
\input{Sections/6_Conclusions}

{\small
\bibliographystyle{ieeetr}
\bibliography{macro,IRISA-IGARSS23}
}

\end{document}

%% file: 0_ieee.tex
\usepackage{spconf,amsmath,epsfig}

\usepackage{cite}

\usepackage{graphicx}
\graphicspath{{Figures/}}
\DeclareGraphicsExtensions{.pdf,.jpeg,.png}

\usepackage{amsmath}
\usepackage{algorithmic}

\usepackage{gensymb}
\usepackage{subfloat}

\usepackage{array}

\usepackage{multirow}
\usepackage{amsfonts}
\usepackage{stfloats}

\usepackage{url}
\usepackage[export]{adjustbox}
\usepackage{textcomp}
\usepackage{cite}

\usepackage{caption}
\usepackage[font=footnotesize]{subcaption}
\usepackage{breqn}

%% file: 0_preamble.tex
\usepackage{lipsum}
\usepackage{framed,multirow}
\usepackage{booktabs}
\usepackage{url}
\usepackage{xcolor}

\usepackage{comment}

\usepackage{algorithm}
\usepackage{eqparbox}

\usepackage{graphicx}
\usepackage{amsmath,amssymb} %
\usepackage{bm}
\usepackage{booktabs}
\usepackage{rotating,multirow}
\usepackage{wrapfig}

\usepackage{tikz}
\usepackage[export]{adjustbox}
\usepackage[inkscapearea=page]{svg}
\setsvg{inkscape=inkscape -z -D,svgpath=fig/}

\graphicspath{{./Images/}{./Figures/result/}{./Images/result/}}

%% file: Sections/0_abstract.tex
Partial multi-task learning where training examples are
annotated for one of the target tasks
is a promising idea in remote sensing
as it allows combining datasets annotated for different tasks
and predicting more tasks with fewer network parameters.
The naïve approach to partial multi-task learning is sub-optimal due
to the lack of all-task annotations for learning joint representations. This paper
proposes using knowledge distillation to replace the need of
ground truths for the alternate task and enhance the performance of such
approach. Experiments conducted on the public ISPRS 2D Semantic Labeling Contest
dataset show the effectiveness of the proposed idea on partial multi-task
learning for semantic tasks including object detection and semantic segmentation in
aerial images. Multi-task learning source codes are available at
\url{https://github.com/lhoangan/multas}.

%% file: Sections/1_Intro.tex
\section{Introduction}
\label{sec:intro}

Remote sensing imagery for earth observation (EO) is an evolving research domain that enables precise recognition and understanding of various substances and objects present on the Earth's surface. Employing deep
learning advancements from the computer vision and deep learning communities improves
performance and accuracy in remote sensing problems but not without hindrances.
The data-demanding nature of deep models requires a large number of annotated
data which are expensive due to the large domain diversities of remote sensing
data, thus limiting its applicability.
The vast number of remote sensing images captured from frequent satellite passes
or aerial acquisitions, however, are not readily usable to train deep networks
developed for generic vision problems due to the lack of task-specific
annotations and possible domain gaps.

Multi-task learning aims to predict different targets from the same inputs, thus
typically requires annotations of all the target tasks for each
input example to learn the interrelationship at the shared encoder
by optimizing all tasks at the same time.
Different from multi-modality where a network is fed with various modalities
and uses complementary information for optimization~\cite{HALe2018}, in
multi-task learning the network is required to extract useful information for
each task from the same input. As it can be observed that remote sensing tasks
are often related, semantically (e.g. building extraction and change
detection~\cite{Hong2023}), geometrically (e.g. land-cover prediction and height
estimation~\cite{gao2023joint}), or structurally (e.g. semantic segmentation and
super-resolution~\cite{Salgueiro2022SEGAN}), etc., successfully exploiting the
interrelationship between tasks could play a critical role in improving the
application of deep learning techniques on remote sensing problems.
Consequently, efforts have been spent to enhance existing datasets
with more task annotations, such as iSAID~\cite{isaid2019} re-annotating the
detection-only DOTA dataset~\cite{dota2018} for semantic and instance segmentation,
ALS2DTM~\cite{Le2022ALS2DTM} provides DTM information to the otherwise point-cloud only
DALES~\cite{dales2020} dataset, etc.

On the other hand, multi-task partially supervised learning, or, in short,
partial multi-task learning~\cite{Li2022MTPSL}
allows each  training example to be annotated for a single task, thus
having the potential for data exploitation as 
combining data sources annotated for different tasks could
improve the members' performances~\cite{Le2023BMVC}.
Similar to traditional multi-task learning, the approach
supposes the complementary relationships of the
tasks and their features' compatibility so that optimizing a single model to
predict various targets from an input would learn better representations.

Despite encouraging performances, due to the lack of mutual annotations of the
inputs, partial multi-task learning optimizes each task alternately, limiting the
ability to learn joint representations and leading to sub-optimal results. In this
paper, we propose using knowledge distillation techniques to provide supervised
signals for training the tasks with unavailable ground truths in the partially supervised
learning paradigm.
In particular, a larger network pre-trained with the
available annotated data is frozen and
provides labels as soft targets for the un-annotated tasks while training the others.
This idea has been mentioned in the original paper~\cite{Le2023BMVC} yet only
the basic feature distillation with mean-square errors is shown.
This paper shows how the usage of soft labels in combination with feature-level
distillation would further improve the performance for remote sensing data.

%% file: Sections/3_Method.tex
\section{Method}
\label{sec:method}

\begin{figure*}[t]
    \centering
    \scriptsize
    \def\svgwidth{.99\textwidth}
    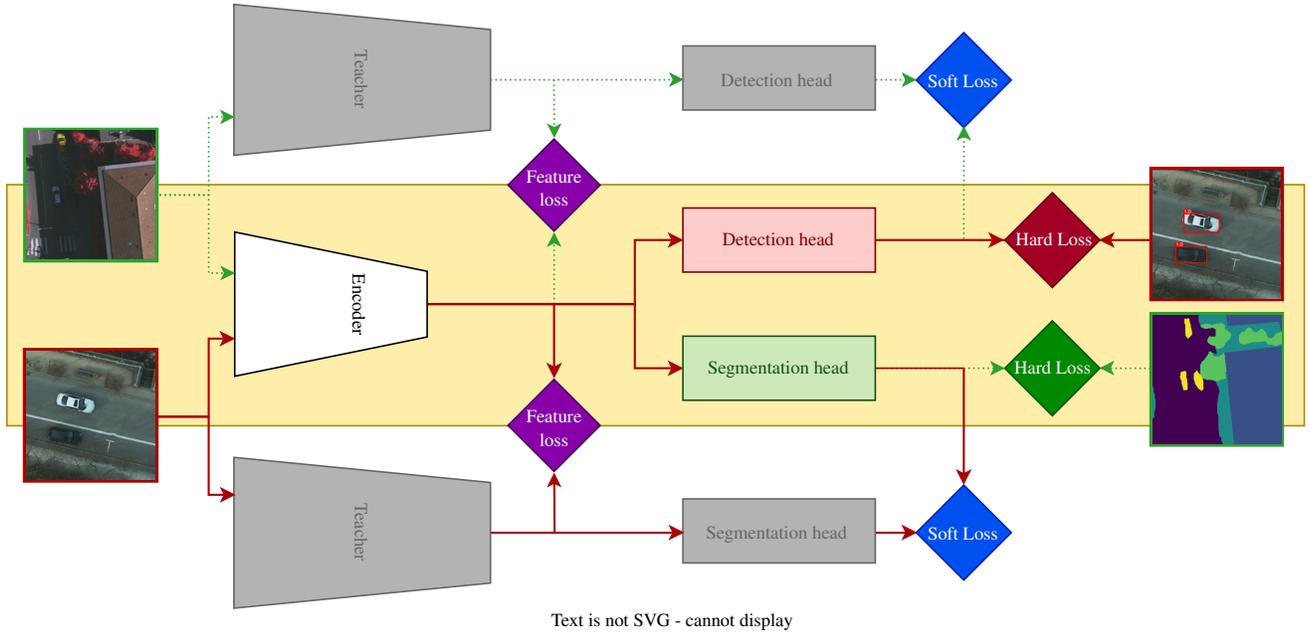
    \caption{Multi-task partially supervised learning with knowledge distillation. 
    The vanilla setup is shown in the {\color[HTML]{aa8800}yellow} box including
    a shared encoder and two task heads for object detection (in {\color[HTML]{a30026} red})
    and semantic segmentation (in {\color[HTML]{008a00} green} with their respective supervised
    losses.
    We illustrate the data flow for the iteration with annotated detection using solid
    {\color[HTML]{aa0000}red} lines while the dotted {\color[HTML]{2ca02c} green} lines
    are for annotated segmentation in the next iteration.
    The teacher networks (in {\color[HTML]{666666}gray}) provide soft labels and/or feature
    loss to train the task head without annotations (illustrated for segmentation).
    }
    \label{fig:illustration}
\end{figure*}

The method in this paper is based on the partial multi-task learning setup by~\cite{Le2023BMVC}
which includes a shared encoder and two output
heads, one for each task. The {\color[HTML]{aa8800}yellow} box in Fig.~\ref{fig:illustration} illustrates the setup
for two tasks, object detection and semantic segmentation, with two optical remote sensing datasets.
An image batch from a detection dataset, bordered with {\color[HTML]{aa0000}red}, is passed through
the network and optimized with the supervised (hard) loss on the detection head while another
from a segmentation dataset, bordered with {\color[HTML]{2ca02c}green}, for the segmentation head.
The object detection loss includes the classification Focal
Loss~\cite{Lin2017focal} and the localization Balanced L1 Loss~\cite{Pang2019} 
while the semantic segmentation head uses the regular cross-entropy with softmax.

As each image is annotated exclusively for only one task,
the two tasks are trained alternately every iteration.
Each task head receives the gradients from 
the respective annotated data while the shared encoder
has the gradients accumulated. The entire network parameters
are only updated once both tasks' data have been passed,
two data passes effectively make up one training iteration.

Although the shared encoder receives the accumulated
gradients from both tasks, the corresponding task head is trained only by those
with available ground truths limiting the
advantage of the extended dataset from the other task.
Inspired by the self-training setup from~\cite{Le2023ICCVw}
where the beneficial supervised signals are provided by a network trained
on a disjoint dataset, a pre-trained teacher is proposed ({\color[HTML]{666666}gray} shapes)
which allows distilling at the logit level, i.e. the soft loss ({\color[HTML]{0050ef}blue} squares
in Figure~\ref{fig:illustration}),
and/or at the feature level, i.e. the feature loss ({\color[HTML]{8800aa}purple} squares).
This teacher network provides training signals for the un-annotated task while
concurrently with the annotated one.

Following the study of knowledge distillation for object detection on earth
observation data~\cite{Le2023knowledge},
two feature-level distillation strategies are employed, namely mean-squared-error
difference and PDF-Distil~\cite{Zhang2021PDF}, coined MSE and PDF, respectively.
For semantic segmentation, only the MSE is used.
MSE minimizes the differences between the teacher's and student's predictions
uniformly across the feature maps while PDF weights the differences based
only teacher-student prediction disagreements. As such, the student is 
encourage to learn the representations for agreed locations and discouraged
if the teacher may not be correct.
The ResNet18 backbone with FPN neck~\cite{Lin2017} and ResNet50 with PAFPN
neck~\cite{Liu2018PAFPN} are used as encoders for the student and teacher
networks respectively, with parameter ratio of 1.6.
As the student feature maps have different shapes
than the teacher's, an adaptive layer with $3\times3$ convolution is employed
before applying the loss function.

%% file: Images/IGARSS-KD-MTL_svg-tex.pdf_tex
\begingroup%
  \makeatletter%
  \providecommand\color[2][]{%
    \errmessage{(Inkscape) Color is used for the text in Inkscape, but the package 'color.sty' is not loaded}%
    \renewcommand\color[2][]{}%
  }%
  \providecommand\transparent[1]{%
    \errmessage{(Inkscape) Transparency is used (non-zero) for the text in Inkscape, but the package 'transparent.sty' is not loaded}%
    \renewcommand\transparent[1]{}%
  }%
  \providecommand\rotatebox[2]{#2}%
  \newcommand*\fsize{\dimexpr\f@size pt\relax}%
  \newcommand*\lineheight[1]{\fontsize{\fsize}{#1\fsize}\selectfont}%
  \ifx\svgwidth\undefined%
    \setlength{\unitlength}{620.04423523bp}%
    \ifx\svgscale\undefined%
      \relax%
    \else%
      \setlength{\unitlength}{\unitlength * \real{\svgscale}}%
    \fi%
  \else%
    \setlength{\unitlength}{\svgwidth}%
  \fi%
  \global\let\svgwidth\undefined%
  \global\let\svgscale\undefined%
  \makeatother%
  \begin{picture}(1,0.50291589)%
    \lineheight{1}%
    \setlength\tabcolsep{0pt}%
    \put(0,0){\includegraphics[width=\unitlength,page=1]{IGARSS-KD-MTL_svg-tex.pdf}}%
    \put(0.2782827,0.41824452){\color[rgb]{0.4,0.4,0.4}\rotatebox{-90}{\makebox(0,0)[t]{\lineheight{1.25}\smash{\begin{tabular}[t]{c}Teacher\end{tabular}}}}}%
    \put(0,0){\includegraphics[width=\unitlength,page=2]{IGARSS-KD-MTL_svg-tex.pdf}}%
    \put(0.27646831,0.24829696){\color[rgb]{0,0,0}\rotatebox{-90}{\makebox(0,0)[t]{\lineheight{1.25}\smash{\begin{tabular}[t]{c}Encoder\end{tabular}}}}}%
    \put(0,0){\includegraphics[width=\unitlength,page=3]{IGARSS-KD-MTL_svg-tex.pdf}}%
    \put(0.59638975,0.41302505){\color[rgb]{0.4,0.4,0.4}\makebox(0,0)[t]{\lineheight{1.25}\smash{\begin{tabular}[t]{c}Detection head\end{tabular}}}}%
    \put(0,0){\includegraphics[width=\unitlength,page=4]{IGARSS-KD-MTL_svg-tex.pdf}}%
    \put(0.59761476,0.29244704){\color[rgb]{0.50196078,0,0}\makebox(0,0)[t]{\lineheight{1.25}\smash{\begin{tabular}[t]{c}Detection head\end{tabular}}}}%
    \put(0,0){\includegraphics[width=\unitlength,page=5]{IGARSS-KD-MTL_svg-tex.pdf}}%
    \put(0.59761476,0.19567975){\color[rgb]{0.08627451,0.31372549,0.08627451}\makebox(0,0)[t]{\lineheight{1.25}\smash{\begin{tabular}[t]{c}Segmentation head\end{tabular}}}}%
    \put(0,0){\includegraphics[width=\unitlength,page=6]{IGARSS-KD-MTL_svg-tex.pdf}}%
    \put(0.42843527,0.34017116){\color[rgb]{1,1,1}\makebox(0,0)[t]{\lineheight{1.25}\smash{\begin{tabular}[t]{c}Feature\\loss\end{tabular}}}}%
    \put(0,0){\includegraphics[width=\unitlength,page=7]{IGARSS-KD-MTL_svg-tex.pdf}}%
    \put(0.73685572,0.0704589){\color[rgb]{1,1,1}\makebox(0,0)[t]{\lineheight{1.25}\smash{\begin{tabular}[t]{c}Soft Loss\end{tabular}}}}%
    \put(0,0){\includegraphics[width=\unitlength,page=8]{IGARSS-KD-MTL_svg-tex.pdf}}%
    \put(0.2782827,0.07628256){\color[rgb]{0.4,0.4,0.4}\rotatebox{-90}{\makebox(0,0)[t]{\lineheight{1.25}\smash{\begin{tabular}[t]{c}Teacher\end{tabular}}}}}%
    \put(0,0){\includegraphics[width=\unitlength,page=9]{IGARSS-KD-MTL_svg-tex.pdf}}%
    \put(0.59638975,0.0710631){\color[rgb]{0.4,0.4,0.4}\makebox(0,0)[t]{\lineheight{1.25}\smash{\begin{tabular}[t]{c}Segmentation head\end{tabular}}}}%
    \put(0,0){\includegraphics[width=\unitlength,page=10]{IGARSS-KD-MTL_svg-tex.pdf}}%
    \put(0.73685572,0.41211169){\color[rgb]{1,1,1}\makebox(0,0)[t]{\lineheight{1.25}\smash{\begin{tabular}[t]{c}Soft Loss\end{tabular}}}}%
    \put(0,0){\includegraphics[width=\unitlength,page=11]{IGARSS-KD-MTL_svg-tex.pdf}}%
    \put(0.80566443,0.29244704){\color[rgb]{1,1,1}\makebox(0,0)[t]{\lineheight{1.25}\smash{\begin{tabular}[t]{c}Hard Loss\end{tabular}}}}%
    \put(0,0){\includegraphics[width=\unitlength,page=12]{IGARSS-KD-MTL_svg-tex.pdf}}%
    \put(0.80449663,0.19567975){\color[rgb]{1,1,1}\makebox(0,0)[t]{\lineheight{1.25}\smash{\begin{tabular}[t]{c}Hard Loss\end{tabular}}}}%
    \put(0,0){\includegraphics[width=\unitlength,page=13]{IGARSS-KD-MTL_svg-tex.pdf}}%
    \put(1.29229421,0.41824452){\color[rgb]{0.4,0.4,0.4}\rotatebox{-90}{\makebox(0,0)[t]{\lineheight{1.25}\smash{\begin{tabular}[t]{c}Teacher\end{tabular}}}}}%
    \put(0,0){\includegraphics[width=\unitlength,page=14]{IGARSS-KD-MTL_svg-tex.pdf}}%
    \put(1.26870718,0.24829696){\color[rgb]{0,0,0}\rotatebox{-90}{\makebox(0,0)[t]{\lineheight{1.25}\smash{\begin{tabular}[t]{c}Student\end{tabular}}}}}%
    \put(0,0){\includegraphics[width=\unitlength,page=15]{IGARSS-KD-MTL_svg-tex.pdf}}%
    \put(1.61040126,0.41302505){\color[rgb]{0.4,0.4,0.4}\makebox(0,0)[t]{\lineheight{1.25}\smash{\begin{tabular}[t]{c}Detection head\end{tabular}}}}%
    \put(0,0){\includegraphics[width=\unitlength,page=16]{IGARSS-KD-MTL_svg-tex.pdf}}%
    \put(1.61162626,0.29244704){\color[rgb]{0.50196078,0,0}\makebox(0,0)[t]{\lineheight{1.25}\smash{\begin{tabular}[t]{c}Detection head\end{tabular}}}}%
    \put(0,0){\includegraphics[width=\unitlength,page=17]{IGARSS-KD-MTL_svg-tex.pdf}}%
    \put(1.61162626,0.19567975){\color[rgb]{0.08627451,0.31372549,0.08627451}\makebox(0,0)[t]{\lineheight{1.25}\smash{\begin{tabular}[t]{c}Segmentation head\end{tabular}}}}%
    \put(0,0){\includegraphics[width=\unitlength,page=18]{IGARSS-KD-MTL_svg-tex.pdf}}%
    \put(1.44228351,0.32873477){\color[rgb]{0,0,0}\makebox(0,0)[t]{\lineheight{1.25}\smash{\begin{tabular}[t]{c}Feature loss\end{tabular}}}}%
    \put(0,0){\includegraphics[width=\unitlength,page=19]{IGARSS-KD-MTL_svg-tex.pdf}}%
    \put(1.75086723,0.0704589){\color[rgb]{1,1,1}\makebox(0,0)[t]{\lineheight{1.25}\smash{\begin{tabular}[t]{c}Soft Loss\end{tabular}}}}%
    \put(0,0){\includegraphics[width=\unitlength,page=20]{IGARSS-KD-MTL_svg-tex.pdf}}%
    \put(1.29229421,0.07628256){\color[rgb]{0.4,0.4,0.4}\rotatebox{-90}{\makebox(0,0)[t]{\lineheight{1.25}\smash{\begin{tabular}[t]{c}Teacher\end{tabular}}}}}%
    \put(0,0){\includegraphics[width=\unitlength,page=21]{IGARSS-KD-MTL_svg-tex.pdf}}%
    \put(1.61040126,0.0710631){\color[rgb]{0.4,0.4,0.4}\makebox(0,0)[t]{\lineheight{1.25}\smash{\begin{tabular}[t]{c}Segmentation head\end{tabular}}}}%
    \put(0,0){\includegraphics[width=\unitlength,page=22]{IGARSS-KD-MTL_svg-tex.pdf}}%
    \put(1.44228351,0.15736021){\color[rgb]{0,0,0}\makebox(0,0)[t]{\lineheight{1.25}\smash{\begin{tabular}[t]{c}Feature loss\end{tabular}}}}%
    \put(0,0){\includegraphics[width=\unitlength,page=23]{IGARSS-KD-MTL_svg-tex.pdf}}%
    \put(1.75086723,0.41211169){\color[rgb]{1,1,1}\makebox(0,0)[t]{\lineheight{1.25}\smash{\begin{tabular}[t]{c}Soft Loss\end{tabular}}}}%
    \put(0,0){\includegraphics[width=\unitlength,page=24]{IGARSS-KD-MTL_svg-tex.pdf}}%
    \put(1.81967594,0.29244704){\color[rgb]{1,1,1}\makebox(0,0)[t]{\lineheight{1.25}\smash{\begin{tabular}[t]{c}Hard Loss\end{tabular}}}}%
    \put(0,0){\includegraphics[width=\unitlength,page=25]{IGARSS-KD-MTL_svg-tex.pdf}}%
    \put(1.81850814,0.19567975){\color[rgb]{1,1,1}\makebox(0,0)[t]{\lineheight{1.25}\smash{\begin{tabular}[t]{c}Hard Loss\end{tabular}}}}%
    \put(0,0){\includegraphics[width=\unitlength,page=26]{IGARSS-KD-MTL_svg-tex.pdf}}%
    \put(0.42843527,0.1587325){\color[rgb]{1,1,1}\makebox(0,0)[t]{\lineheight{1.25}\smash{\begin{tabular}[t]{c}Feature\\loss\end{tabular}}}}%
    \put(0.51761581,0.00544316){\makebox(0,0)[t]{\lineheight{1.25}\smash{\begin{tabular}[t]{c}Text is not SVG - cannot display\end{tabular}}}}%
  \end{picture}%
\endgroup%

%% file: Sections/5_Experiments.tex
\section{Experiments}
\label{sec:exp}

\subsection{Dataset and setup}

\begin{figure*}[t]
    \centering
    \includegraphics[width=.40\textwidth]{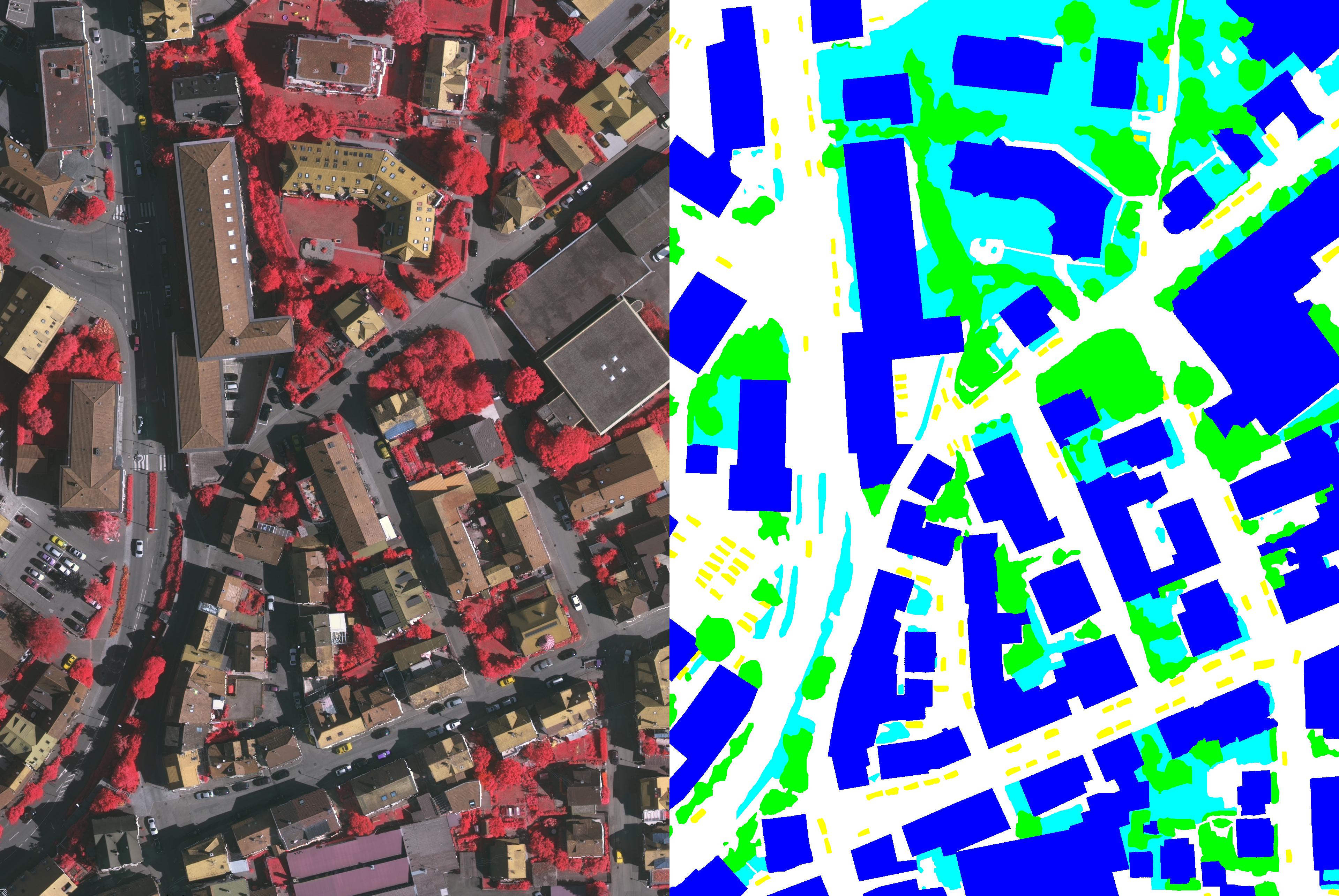}
    \includegraphics[width=.40\textwidth]{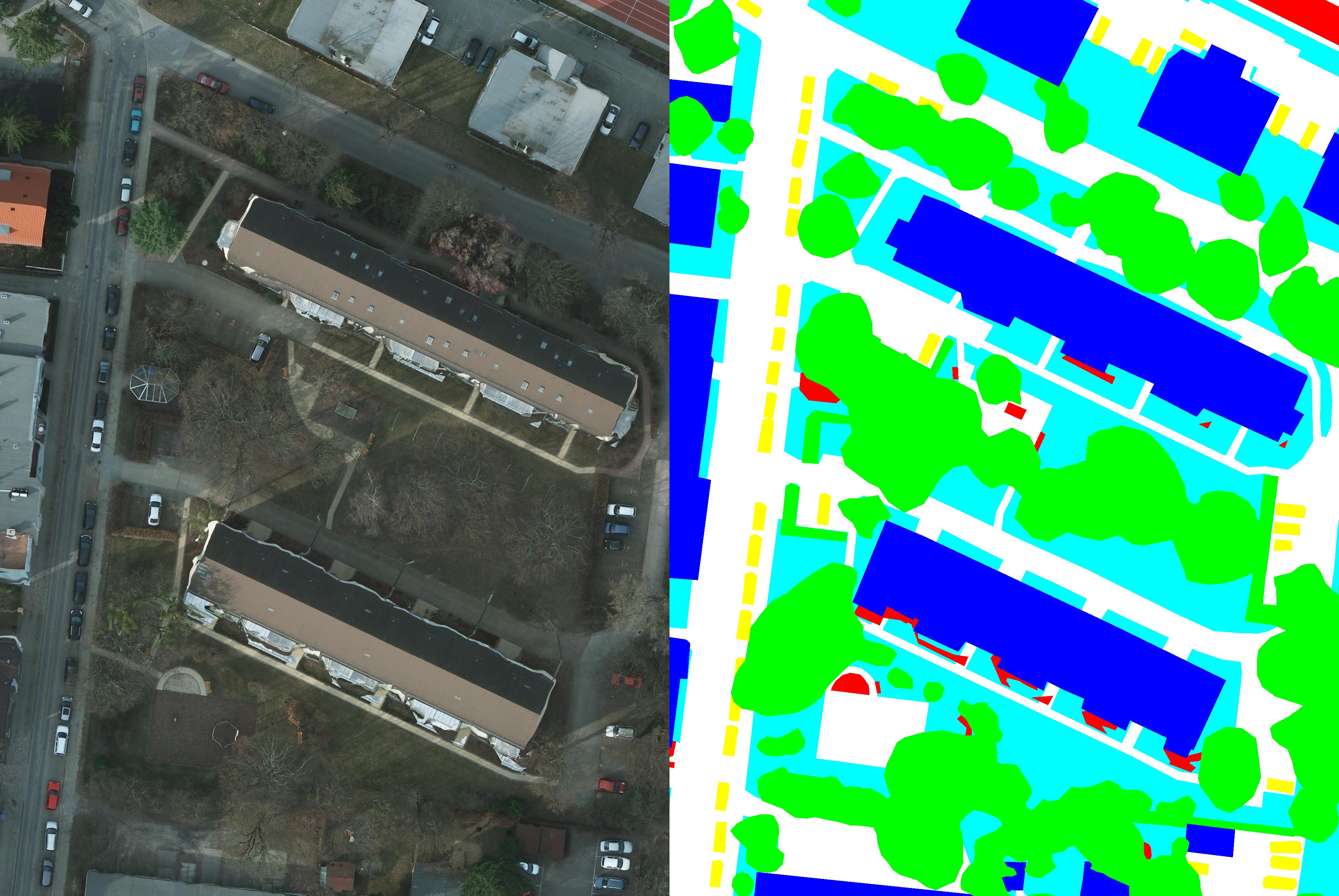}
    \caption{A sample of Vaihingen (left) and Postdam (right) subsets~\cite{isprs2012}.
    Vaihingen data compose of IR-R-G channels thus appear with reddish vegetation while Potsdam is with 
    regular R-G-B imagery. The lower ground sampling distance of Vaihingen results in smaller and
    more objects in the same image crop compared to Potsdam.
    }
    \label{fig:vaihingen-potsdam}
\end{figure*}

The ISPRS 2D semantic labeling contest~\cite{isprs2012} is employed for experimentation.
The dataset composes of aerial data collected from two different cities, Vaihingen and 
Potsdam with typical differences found in remote sensing datasets.
Vaihingen contains 33 IR-R-G tiles of around 3K$\times$2K resolution with ground
sampling distance (GSD) of 9cm whereas Potsdam contains 38 R-G-B tiles of 6K$\times$6K
resolution with GSD of 5cm. The lower GSD results in more pixels for the same
object types, e.g. cars, for Potsdam while the different sensor composition
creates reddish vegetation for Vaihingen with different statistical
distributions in each channels. In addition, the two subsets are divergent
in scene appearance due to geographical differences: small and detached houses
in Vaihingen and complex settlement structure with narrow streets in Potsdam.
The tiles are cropped into $320\times320$-pixel image chips. The chips are, then,
split into training and validation, resulting in 
632/702 for Vaihingen and 2993/1515 Potsdam. 

\textbf{Semantic segmentation.} The ISPRS benchmark comes with pixel-wise annotations for urban land cover,
including 6 categories: \emph{impervious surfaces, building, low vegetation,
tree, car, and background}. The mean intersection-over-union (mIoU) metric is reported.
Evaluation is performed on the eroded masks using a 3-pixel radius circular disc
to reduce the impact of uncertain border definitions on the evaluation~\cite{rottensteiner2012isprs}.

\textbf{Object detection.} We derive bounding-box targets from the provided pixel-wise semantic labels
by taking a blob of connected pixels of the same class as an instance.
To that end, the \emph{building}, \emph{tree}, and \emph{car} class are selected for their interest in
urban object detection and the performance is measured using the
VOC mAP@[.5:.95:.05]~\cite{PascalVOC} metric.

\subsection{Results}

We evaluate the benefit of knowledge distillation with partial multi-task
learning with Potsdam set used for object detection and Vaihingen for semantic
segmentation in Table~\ref{tab:P+V} and vice versa in Table~\ref{tab:V+P}. The
teachers' performances are also included for reference.

\input{Tables/Table1}

It can be seen that on remote sensing data, partial multi-task learning does not
gain much benefit over single-task learning: the results are slightly higher for
semantic segmentation and on par or slightly lower for object detection. This
could be attributed to the differences in domains between Potsdam and Vaihingen
data. Vaihingen contains IR-R-G images with ground sampling distance (GSD) of 9cm 
whereas Potsdam contains R-G-B images with GSD of 5cm. Thus, same object types
would appear larger in Potsdam while vegetation appears reddish in Vaihingen, in
addition to geographical differences.

By leveraging the use of knowledge distillation, adding soft labels in place of unavailable ground truths or feature
distillation is generally helpful. This agrees to the discovery found in~\cite{Le2023BMVC}
although they only examine with MSE setting. Combining soft label with feature
distillation, especially PDF-Distil results in the highest performance.

%% file: Tables/Table1.tex
\begin{table}[t]
    \centering
    \setlength{\tabcolsep}{2pt}
    \setlength\doublerulesep{0.8pt} %
    \begin{tabular}{@{}lcc@{}} %
        \toprule
        Training     & Detection - Postdam \\
        \midrule       
        Teacher      &     51.90        \\ %
        \midrule       
        Single-task  &     45.37        \\ %
        Multi-task   &     45.33        \\ %
        + Soft       &     46.23        \\ %
        + MSE        &     45.47        \\ %
        + PDF        &     46.49        \\ %
        + Soft + MSE &     46.34        \\ %
        + Soft + PDF & \bf 46.63        \\ %
        \bottomrule
    \end{tabular}~
    \begin{tabular}{@{}lc@{}}
        \toprule
                    & Segmentation - Vaihingen\\
        \midrule
                    &        70.96     \\ %
        \midrule
                    &        66.15     \\ %
                    &        68.60     \\ %
                    &        68.63     \\ %
                    &        68.29     \\ %
                    &        67.99     \\ %
                    &        68.60     \\ %
                    &  \bf   68.97     \\ %
        \bottomrule
    \end{tabular}
    \caption{Performance of partial multi-task learning with knowledge distillation with object
    detection on Potsdam set and semantic segmentation task on Vaihingen. Without knowledge
    distillation, (partial) multi-task learning is on par with single task for object detection
    and slightly better for semantic segmentation. The distances are more pronounced when
    knowledge distillation is employed.
    }
    \label{tab:P+V}
\end{table}

\begin{table}[t]
    \centering
    \setlength{\tabcolsep}{2pt}
    \setlength\doublerulesep{0.8pt} %
    \begin{tabular}{@{}lcc@{}} %
        \toprule
        Training     & Detection - Vaihingen \\
        \midrule       
        Teacher      &         48.16        \\ %
        \midrule       
        Single task  &         45.07        \\ %
        Multi-task   &         44.13        \\ %
        + Soft       &         44.83        \\ %
        + MSE        &         44.44        \\ %
        + PDF        &         45.22        \\ %
        + Soft + MSE &         44.61        \\ %
        + Soft + PDF &  \bf    45.49        \\ %
        \bottomrule
    \end{tabular}~
    \begin{tabular}{@{}lc@{}}
        \toprule
                    & Segmentation - Potsdam\\
        \midrule
                    &             66.60     \\ %
        \midrule
                    &             59.24     \\ %
                    &             60.56     \\ %
                    &             61.60     \\ %
                    &             60.93     \\ %
                    &   \bf       61.76     \\ %
                    &             61.70     \\ %
                    &             61.66     \\ %
        \bottomrule
    \end{tabular}
    \caption{
    Performance of partial multi-task learning with knowledge distillation with object
    detection on Vaihingen set and semantic segmentation task on Potsdam. Without knowledge
    distillation, (partial) multi-task learning is on par with single task for object detection
    and slightly better for semantic segmentation. The distances are more pronounced when
    knowledge distillation is employed.
    }
    \label{tab:V+P}
\end{table}

%% file: Sections/6_Conclusions.tex
\section{Conclusion}
\label{sec:conclusion}

This paper studies the multi-task partially supervised learning paradigm for
remote sensing datasets and proposes using knowledge distillation to provide
training signals for non-annotated tasks.
Partial multi-task learning has the potential in remote sensing because of
its efficiency in training as it combines different data sources
from different tasks to expand the training size and allows to perform
various predictions at the cost of fewer network parameters.
The experimental results on the ISPRS urban object detection and segmentation
show that the additional soft and feature loss is generally helpful
and combining them could further improve the performance of both tasks.